\documentclass[sn-mathphys-num]{sn-jnl}
\usepackage{pifont}
\usepackage{booktabs}
\usepackage{graphicx}%
\usepackage{multirow}%
\usepackage{amsmath,amssymb,amsfonts}%
\usepackage{amsthm}%
\usepackage{mathrsfs}%
\usepackage[title]{appendix}%
\usepackage{xcolor}%
\usepackage{ragged2e}
\usepackage{tabularx}
\usepackage{pifont}
\usepackage{makecell}
\usepackage{booktabs}
\usepackage{blindtext}
\usepackage{textcomp}%
\usepackage{manyfoot}%
\usepackage{booktabs}%
\usepackage{pifont}
\usepackage{algorithm}%
\usepackage{algorithmicx}%
\usepackage{algpseudocode}%
\usepackage{listings}%
\usepackage{caption}
\usepackage{subcaption}
\usepackage{booktabs}
\usepackage{tabularx}
\usepackage{siunitx} 
\usepackage{mhchem} 
\usepackage[nameinlink,noabbrev]{cleveref}
\usepackage[export]{adjustbox}
\usepackage{changes}



\begin{document}

\title[Article Title]{Vision Transformers for Preoperative CT-Based Prediction of Histopathologic Chemotherapy Response Score in High-Grade Serous Ovarian Carcinoma}

\author*[1,2]{\fnm{Francesca} \sur{Fati}}
\equalcont{These authors contributed equally to this work.}
\author*[3]{\fnm{Felipe} \sur{ Coutinho}}
\equalcont{These authors contributed equally to this work.}
\author[4,5,6,7]{\fnm{Marika} \sur{Reinius}}
\author[1]{\fnm{Marina} \sur{Rosanu}}
\author[4,5,6,7]{\fnm{Gabriel} \sur{Funingana}}
\author[8]{\fnm{Luigi} \sur{De Vitis}}
\author[1]{\fnm{Gabriella} \sur{Schivardi}}
\author[4,5,6,9]{\fnm{Hannah} \sur{Clayton}}
\author[1]{\fnm{Alice} \sur{Traversa}}
\author[4,5,6,9]{\fnm{Zeyu} \sur{Gao}}
\author[2]{\fnm{Guilherme} \sur{Penteado}}
\author[4,5,6,9]{\fnm{Shangqi} \sur{Gao}}
\author[2]{\fnm{Francesco} \sur{Pastori}}
\author[10]{\fnm{Ramona} \sur{Woitek}}
\author[11]{\fnm{Maria Cristina} \sur{Ghioni}}
\author[1,12]{\fnm{Giovanni Damiano} \sur{Aletti}}
\author[5,7]{\fnm{Mercedes} \sur{Jimenez-Linan}}
\author[5,6]{\fnm{Sarah} \sur{Burge}}
\author[1]{\fnm{Nicoletta} \sur{Colombo}}
\author[13]{\fnm{Evis} \sur{Sala}}
\author[14]{\fnm{Maria Francesca} \sur{Spadea}}
\author[15]{\fnm{Timothy L.} \sur{Kline}}
\author[4,5,6,7]{\fnm{James D.} \sur{Brenton}}
\author[3]{\fnm{Jaime} \sur{Cardoso}}
\author[1,12]{\fnm{Francesco} \sur{Multinu}}
\equalcont{These authors contributed equally to this work.}
\author[1,2]{\fnm{Elena} \sur{De Momi}}
\equalcont{These authors contributed equally to this work.}
\author[4,5,6,9]{\fnm{Mireia} \sur{Crispin-Ortuzar}}
\equalcont{These authors contributed equally to this work.}
\author[4,5,6,9]{\fnm{Ines} \sur{P. Machado}}
\equalcont{These authors contributed equally to this work.}
\affil[1]{%
  \centering
  \small
  Department of Gynecologic Oncology, European Institute of Oncology, IEO, IRCCS, Italy}
\affil[2]{%
  \centering
  \small
  Department of Electronics, Information and Bioengineering, Politecnico di Milano, Italy}
\affil[3]{%
  \centering
  \small
  INESC TEC, Faculty of Engineering, University of Porto, Portugal}
\affil[4]{%
  \centering
  \small
  Department of Oncology, University of Cambridge, United Kingdom}
\affil[5]{%
  \centering
  \small
  Cancer Research UK Cambridge Centre, United Kingdom}
\affil[6]{%
  \centering
  \small
  Cancer Research UK Cambridge Institute, United Kingdom}
\affil[7]{%
  \centering
  \small
  Cambridge University Hospitals NHS Foundation Trust, United Kingdom}
\affil[8]{%
  \centering
  \small
  Department of Obstetrics and Gynecology, Mayo Clinic, USA}
\affil[9]{%
  \centering
  \small
  Early Cancer Institute, University of Cambridge, United Kingdom}
\affil[10]{%
  \centering
  \small
  Research Center for Medical Image Analysis and AI, Danube Private University, Austria}
\affil[11]{%
  \centering
  \small
  Department of Pathology, European Institute of Oncology, IEO, IRCCS, Italy}
\affil[12]{%
  \centering
  \small
  Department of Oncology and Hemato-Oncology, University of Milan, Italy}
\affil[13]{%
  \centering
  \small
  Department of Radiologic Sciences, Università Cattolica del Sacro Cuore, Italy}
\affil[14]{%
  \centering
  \small
  Institute of Biomedical Engineering, Karlsruhe Institute of Technology, Germany}
\affil[15]{%
  \centering
  \small
  Department of Radiology, Mayo Clinic, USA}

\abstract{\textbf{Purpose.} High-grade serous ovarian carcinoma (HGSOC) is characterized by pronounced biological and spatial heterogeneity and is frequently diagnosed at an advanced stage. Neoadjuvant chemotherapy (NACT) followed by delayed primary surgery is commonly employed in patients unsuitable for primary cytoreduction. The Chemotherapy Response Score (CRS) is a validated histopathological biomarker of response to NACT, but it is only available postoperatively. In this study, we investigate whether pre-treatment computed tomography (CT) imaging and clinical data can be used to predict CRS as an investigational decision-support adjunct to inform multidisciplinary team (MDT) discussions regarding expected treatment response. \textbf{Methods.} We proposed a 2.5D multimodal deep learning framework that processes lesion-dense omental slices using a pre-trained Vision Transformer encoder and integrates the resulting visual representations with clinical variables through an intermediate fusion module to predict CRS.
\textbf{Results.} Our multimodal model, integrating imaging and clinical data, achieved a ROC-AUC of 0.95 alongside 95\% accuracy and 80\% precision on the internal test cohort (IEO, n=41 patients). On the external test set (OV04, n=70 patients), it achieved a ROC-AUC of 0.68, alongside 67\% accuracy and 75\% precision. \textbf{Conclusion.} These preliminary results demonstrate the feasibility of transformer-based deep learning for preoperative prediction of CRS in HGSOC using routine clinical data and CT imaging. As an investigational, pre-treatment decision-support tool, this approach may assist MDT discussions by providing early, non-invasive estimates of treatment response.}


\maketitle

\section{Introduction}\label{sec:introduction}

Despite significant progress in treatment, ovarian cancer remains one of the leading causes of gynecological cancer-related deaths among women \cite{Vergote2010}. High-grade serous ovarian carcinoma (HGSOC) is the most prevalent subtype and is characterized by considerable heterogeneity, often presenting as advanced, multi-site metastatic disease \cite{crispin2023integrated, machado2024self}. For patients with advanced HGSOC, treatment strategies include primary cytoreductive surgery (PCS) followed by chemotherapy, or neoadjuvant chemotherapy (NACT) followed by delayed primary surgery (IDS) \cite{Kehoe2015, drury2025multi}. According to recent ASCO guidelines \cite{gaillard2025neoadjuvant}, PCS is preferred for patients who are fit for surgery and in whom complete cytoreduction is considered achievable, whereas NACT followed by IDS is recommended when complete cytoreduction is unlikely or when perioperative risk is high. The goal of NACT is to reduce tumor burden preoperatively, thereby improving the feasibility of subsequent surgical resection and increasing the likelihood of achieving optimal cytoreduction. This approach is particularly important for patients in whom initial debulking surgery is not feasible due to unresectable disease or comorbidities. 

Introduced in 2015, the Chemotherapy Response Score (CRS) is a key histopathological tool for assessing response to NACT, stratifying patients into three categories: CRS1 (minimal or no response), CRS2 (partial response), and CRS3 (complete or near-complete response) \cite{bohm2015chemotherapy}. The CRS system assesses the extent of tumor regression, focusing on changes in tumor cells and the surrounding stroma \cite{zannoni2025chemotherapy}. CRS1 and CRS2 correlate with poorer outcomes whereas CRS3 is associated with improved progression-free survival (PFS) and overall survival (OS) \cite{lee2017external,rajkumar2018prognostic}. Validated across multiple clinical cohorts, CRS has proven valuable not only as a prognostic tool but also as a tool for guiding maintenance treatment decisions \cite{colombo2023consensus}. Despite its clinical relevance, CRS assessment requires the presence of residual disease in the omentum and can only be performed postoperatively, limiting its applicability in some patients treated with NACT. These limitations highlight the need for complementary, non-invasive approaches capable of assessing treatment response earlier in the disease course. In this context, computed tomography (CT), which is routinely used for disease evaluation, offers a readily available alternative and has demonstrated predictive value for PFS and OS. Building on this rationale, previous work from our research group showed that pre- and post-NACT volumetric analysis of omental deposits in HGSOC can predict CRS \cite{rundo2022clinically}. Non-invasive predictors of likely chemosensitivity could support MDT discussions in borderline cases by providing additional context on the expected benefit of an NACT-first strategy versus primary debulking surgery. In particular, a predicted good-response phenotype may increase confidence in selecting NACT when the likelihood of complete upfront cytoreduction is uncertain, whereas a predicted poor-response phenotype may prompt closer scrutiny of whether primary debulking surgery should be favored when feasible. Here, we propose the first deep learning–based method to predict CRS scores from pre-treatment CT scans and clinical data of HGSOC patients, providing a foundation for further research into non-invasive assessment of treatment response and patient stratification.

\section{Methods}
\subsection{Model architecture}
\begin{figure}
    \centering
    \includegraphics[width=\linewidth]{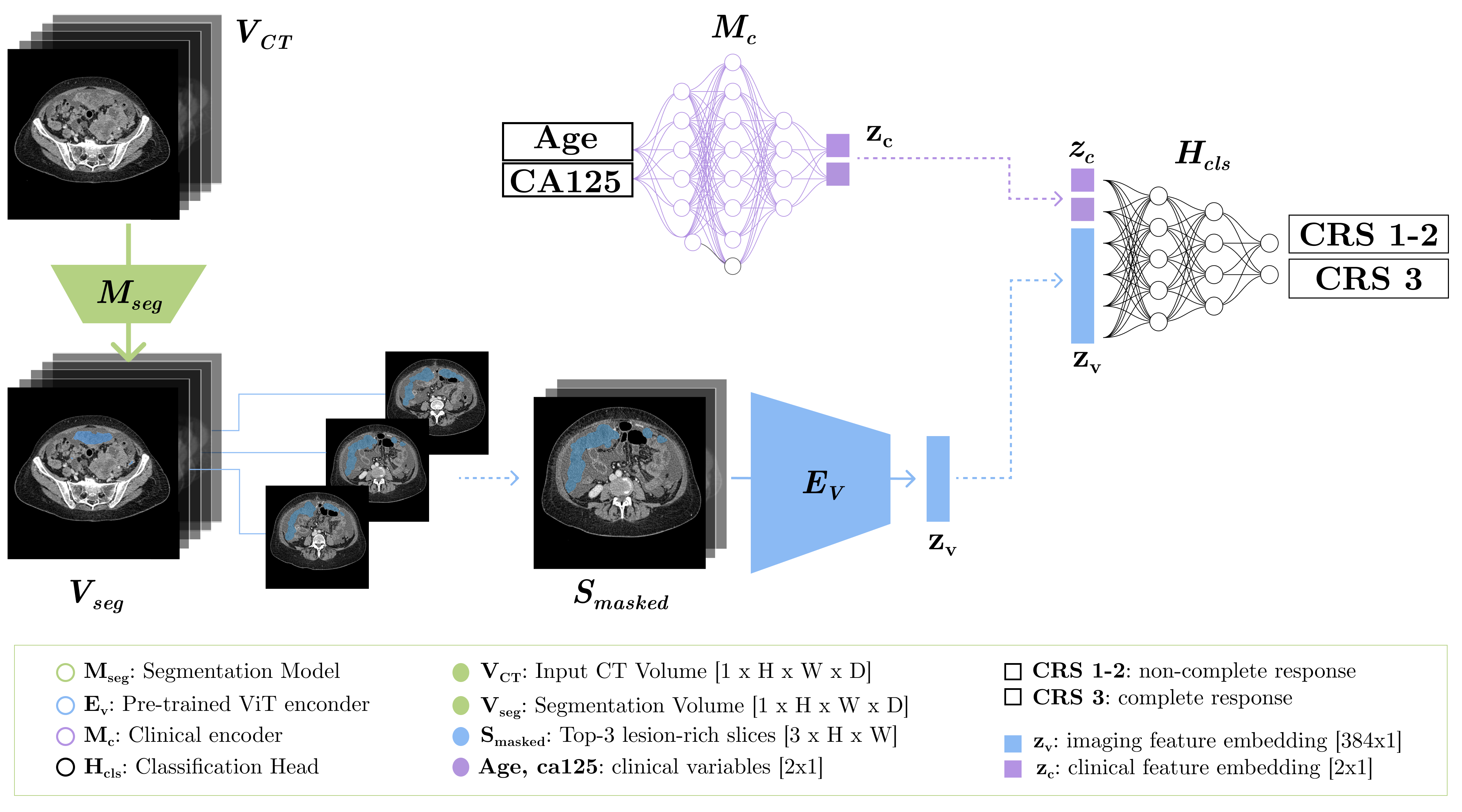}
    \caption{Schematic of the proposed 2.5D multimodal classification pipeline. The three axial slices with the highest lesion density are extracted from the segmented omentum and stacked to form a multi-channel input for a DINOv3-small encoder. The resulting visual representation is combined with encoded clinical variables (age, CA125) via an intermediate fusion layer. The final MLP head outputs the probability of a non- or partial pathological chemotherapy response (CRS 1–2) versus a complete response (CRS 3).}
    \label{fig:model_architecture}
\end{figure}

The proposed pipeline, shown in Figure~\ref{fig:model_architecture}, builds on the multimodal architecture introduced in previous work~\cite{fati2025deep} and is adapted here for binary CRS classification. Specifically, the model distinguishes non- or partial responders (CRS 1–2) from patients achieving a complete response (CRS 3). Our approach, denoted as $M$, combines features extracted from pre-treatment CT scans, $\mathbf{V}_{\text{CT}} \in \mathbb{R}^{H \times W \times D \times 1}$, with a vector of $n$ clinical non-imaging features, $\mathbf{c} \in \mathbb{R}^n$, and outputs a binary prediction $y$,

\begin{equation}
    y = M(\mathbf{V}_{\text{CT}}, \mathbf{c}),
\end{equation}
\newline
where $H$, $W$ and $D$ denote height, width and depth dimensions. 

The pre-treatment CT volumes, $\mathbf{V}_{\text{CT}}$, were rescaled to an isotropic resolution of $1 \times 1 \times 1$ mm, resized to $224 \times 224 \times 128$ voxels, and processed using standard soft-tissue windowing. Omental lesions were segmented using a segmentation model, $M_S$:

\begin{equation}
\mathbf{V}_{\text{seg}} = M_S(\mathbf{V}_{\text{CT}}), \quad \mathbf{V}_{\text{seg}} \in \mathbb{R}^{H \times W \times D \times 1}.
\end{equation} 
\newline
A masked volume was generated via element-wise multiplication, $\mathbf{V}_{\text{masked}} = \mathbf{V}_{\text{CT}} \odot \mathbf{V}_{\text{seg}}$. From $\mathbf{V}_{\text{masked}}$, the three axial slices containing the highest density of lesion-labeled pixels were selected. Preserving their original cranio–caudal anatomical order, the slices were concatenated along the channel dimension to create a 2.5D multi-channel input, $\mathbf{S}_{\text{masked}} \in \mathbb{R}^{H \times W \times 3}$, where $H = W = 224$.

The stacked input, $\mathbf{S}_{\text{masked}}$, is processed using a pre-trained DINOv3-small Vision Transformer (ViT) encoder, $\mathbf{E}{\text{v}}$\footnote{\url{https://github.com/facebookresearch/dinov3}}. This approach allows the encoder to capture spatial relationships across the most informative lesion slices while maintaining their anatomical order. Although DINOv3 was pre-trained on large-scale natural image datasets outside the medical imaging domain, it provides strong visual representations that facilitate the extraction of discriminative imaging features. Finally, the transformer classification token ([CLS]) is extracted to serve as a compact, volume-level representation:

\begin{equation}
\mathbf{z}_{\text{V}} = \mathbf{E}_{\text{v}}(\mathbf{S}_{\text{masked}}), \quad \mathbf{z}_{\text{V}} \in \mathbb{R}^{384}.
\end{equation}\\

Simultaneously, the clinical feature vector, $\mathbf{c}$, is encoded by a clinical variable encoding module, $M_{c}$, into a learned representation, $\mathbf{z}_{\text{clin}} \in \mathbb{R}^n$, as
\begin{equation}
    \mathbf{z}_{\text{clin}} = M_c(\mathbf{c}),
\quad
\mathbf{z}_{\text{clin}} \in \mathbb{R}^{n}.
\end{equation}

Finally, the learned CT volume embedding and the clinical feature embeddings are combined through an intermediate fusion operation and passed to an MLP-based classification head, $H_{\text{cls}}$, which includes ReLU intermediate activations and batch normalization modules:

\begin{equation}
\hat{p} = H_{\text{cls}}\!\left(
\left[\mathbf{z}_{\text{V}} \, \| \, \mathbf{z}_{\text{clin}}\right]
\right),
\quad
\hat{p} \in [0,1],
\end{equation} \newline

where $\hat{p}$ represents the predicted probability of a non- or partial pathological chemotherapy response (CRS 1–2) versus a complete response (CRS 3). The corresponding class label, $\hat{y} \in {0,1}$, is then determined by applying a decision threshold.

\subsection{Training strategy}
A hybrid transfer learning framework was employed, in which the pre-trained ViT encoder was frozen, and a task-specific classification head was fine-tuned for binary chemotherapy response prediction. Training used the AdamW optimizer ($\beta_1 = 0.9$, $\beta_2 = 0.999$, $\epsilon = 10^{-8}$, weight decay $= 10^{-7}$) for up to 200 epochs with early stopping. A batch size of 42 CT volumes and a dropout rate of 0.25 were applied. The learning rate followed a linear warm-up to $10^{-6}$, followed by linear decay. The model was trained using a weighted binary cross-entropy (WBCE) loss. A weighted random sampler was employed during training to balance class representation within each batch.

\section{Experiments}

\subsection{Patient cohorts}
The model was trained and internally evaluated using a cohort of patients treated at the European Institute of Oncology, Milan (“IEO”, n = 271) between January 2016 and December 2023. The study was approved by the IEO Scientific Board (UID 4134), and all patients provided informed consent for the use of their data for research purposes. Additionally, a prospective observational study from Addenbrooke’s Hospital, Cambridge University Hospitals (“OV04”, n = 70) was included as an independent external validation set. OV04 patients were treated at Cambridge University Hospitals NHS Foundation Trust between 2009 and 2020 and were recruited into a prospective clinical study approved by the local research ethics committee (REC reference numbers: 08/H0306/61). For both datasets, patients had a confirmed histopathological diagnosis of HGSOC and were treated with neoadjuvant chemotherapy prior to delayed primary surgery.

\subsection{Image acquisition and labelling}
\begin{figure}[ht]
    \centering
    \includegraphics[width=1\linewidth]{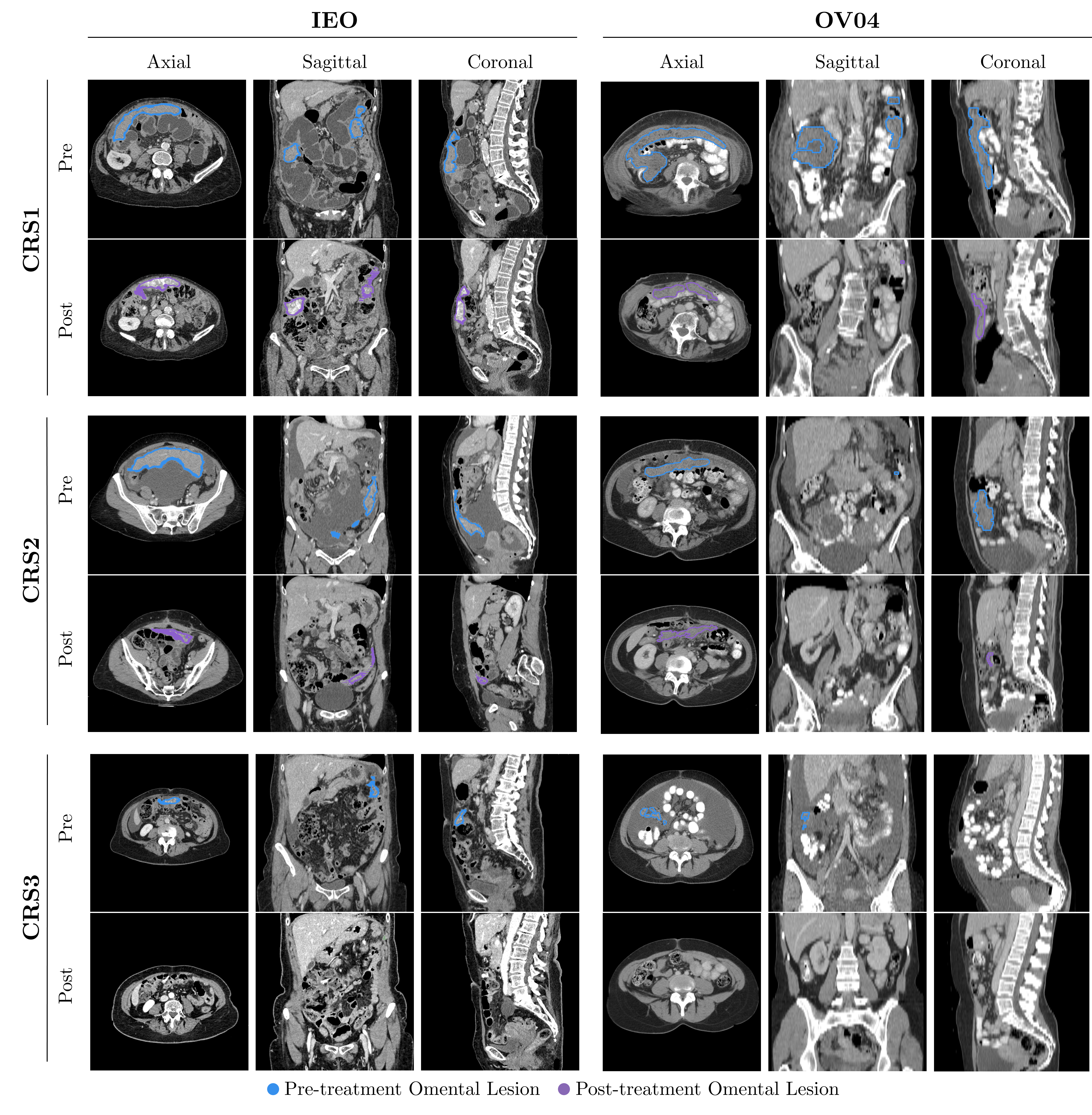}
    \caption{Representative pre- and post-treatment CT scans of HGSOC patients from the IEO and OV04 cohorts, stratified by CRS grade: CRS 1 (minimal or no response), CRS 2 (partial response), and CRS 3 (complete or near-complete response). Axial, sagittal, and coronal views are shown for each case, with omental tumor regions used for CRS assessment highlighted in blue on pre-treatment scans and in violet on post-treatment scans.}
    \label{fig:samples}
\end{figure}

Clinical contrast-enhanced CT (CE-CT) scans covering the abdomen and pelvis were acquired. On CE-CT axial images reconstructed with a slice thickness of 5 mm, all cancer lesions were segmented using OvSeg \cite{buddenkotte2023deep}, a deep learning–based segmentation network for metastatic ovarian cancer. For the OV04 external cohort, cancer lesions were segmented by a board-certified radiologist with ten years of experience in clinical imaging using Microsoft Radiomics (project InnerEye; Microsoft, Redmond, WA, USA). Pre- and post-treatment CT scans used for CRS assessment are illustrated in Figure~\ref{fig:samples}, with omental tumor regions highlighted for each patient. 

\subsection{Clinical data}
Two clinical features were incorporated: patient age and cancer antigen 125 (CA-125), an established serum tumor marker in epithelial ovarian cancer, with concentrations (U/mL) measured at the time of diagnosis. These features were chosen based on their prognostic value in ovarian cancer and their routine availability in clinical practice. Baseline clinical characteristics including patient age, CA-125 levels, and CRS distribution for the IEO and OV04 cohorts are summarized in Table~\ref{tab:cohort_characteristics}.

\begin{table}[ht]
\centering
\caption{Baseline clinical characteristics and chemotherapy response score (CRS) distribution of the IEO and OV04 cohorts. The table reports patient age and serum cancer antigen 125 (CA-125) levels at diagnosis, together with CRS labels, for the IEO cohort (subdivided into training, validation, and internal test sets) and the independent external OV04 cohort. Continuous variables are presented as median [IQR].}
\label{tab:cohort_characteristics}
\begin{tabularx}{\linewidth}{lXXXX}
\toprule
\textbf{Characteristic} & \multicolumn{3}{c}{\makecell{\textbf{IEO}\\\textbf{(Internal)}}} &
\makecell{\textbf{OV04}\\\textbf{(External)}} \\
\cmidrule(lr){2-4}
& \makecell{Train \\ (n=191)} & \makecell{Validation \\ (n=39)} & \makecell{Test \\(n=41)} & \makecell{ Test \\ (n=70)} \\
\midrule
Age (years) & 64 \tiny{[54--70]} & 62 \tiny{[55--66]} & 65 \tiny{[55--74]} & 63 \tiny{[54--70]} \\
CA-125 (U/mL) & 1192 \tiny{[404--2896]} & 1006 \tiny{[345--1722]} & 914 \tiny{[472--2875]} & 898 \tiny{[286--2070]} \\
\midrule
CRS, n (\%) & & & & \\
\quad CRS 1 (no/minimal) & 80 (41.7\%) & 17 (43.6\%) & 17 (41.5\%) & 19 (27.1\%) \\
\quad CRS 2 (partial) & 86 (44.8\%) & 18 (46.1\%) & 18 (43.9\%) & 21 (30.0\%) \\
\quad CRS 3 (complete) & 26 (13.5\%) & 6 (15.4\%) & 6 (14.6\%) & 30 (42.9\%) \\
\bottomrule
\end{tabularx}
\end{table}

\subsection{Histopathologic analysis}
The three-tier CRS was assigned by board-certified pathologists with subspecialty training in gynecological oncology at both centers, following previously published criteria and ICCR recommendations \cite{bohm2015chemotherapy,bohm2019histopathologic}. Briefly, the section of omentum showing the greatest amount of residual viable tumor was assigned a score based on the histopathologic response to chemotherapy: score 1 = abundant tumor with no or minimal perceptible response; score 2 = a significant amount of viable tumor, showing a readily appreciable fibro-inflammatory response secondary to treatment; score 3 = complete or near-complete response, with no tumor or only minimal, irregularly scattered tumor nests (no tumor deposits $>$ 2 mm). For all model-fitting analyses, the three-tier CRS outcomes were dichotomized into non-complete response (CRS 1–2) and complete response (CRS 3). Table \ref{tab:baseline_characteristics} reports the median tumor volume, surface area, and largest connected component fraction for each CRS grade, enabling comparison between the internal (IEO) and external (OV04) cohorts.

\begin{table}[t]
\centering
\caption{Comparison of patient demographics and omental tumor morphology across CRS grades in the IEO internal validation set and the OV04 external validation set. Data are presented as median values for age, pre-treatment CA-125 levels, and quantitative tumor morphology metrics, including tumor volume, surface area, and the largest connected component (CC) fraction. Patients are stratified by CRS classification: CRS 1 (minimal or no response), CRS 2 (partial response), and CRS 3 (complete or near-complete response).}

\label{tab:baseline_characteristics}
\begin{tabularx}{\linewidth}{lXXXXXX}
\toprule
\textbf{} & \multicolumn{3}{c}{\makecell{\textbf{IEO}\\\textbf{(Internal)}}} &
\multicolumn{3}{c}{\makecell{\textbf{OV04}\\\textbf{(External)}}}\\
\cmidrule(lr){2-4} \cmidrule(lr){5-7}
& \makecell{CRS1 \\ (n=114)} & \makecell{CRS2\\ (n=122)} & \makecell{CRS3 \\(n=38)} & \makecell{CRS1 \\ (n=19)} & \makecell{CRS2\\ (n=21)} & \makecell{CRS3 \\(n=30)} \\
\midrule
Age (years) & 65.5 & 63.5 & 60.5 & 64.2 & 65.5 & 61.0 \\
CA-125 (U/mL) & 655.9 & 946.7 & 426.5 & 867.0 & 584.0 & 1423.5 \\
\midrule
Tumor Morphology & & & & \\
\quad Volume [$cm^3$] & 102.4 & 80.98  & 15.62 & 119.2 & 61.04  & 36.6 \\
\quad Surface Area [$cm^2$] &468.6 & 437.8 & 167.4 & 1276 &  781 & 457.9\\
\quad Largest CC Fraction & 0.91  & 0.87 & 0.73&  0.91 & 0.85 & 0.69\\
\midrule 
Therapy Cycles & & & & \\
\quad 3-4 cycles & 114 & 122 & 38 & 13 & 18 & 26 \\ 
\quad +5 cycles & - & - & - & 6 & 3 & 4 \\
\bottomrule
\end{tabularx}
\end{table}

\newpage
\subsection{Performance metrics}
We evaluated response prediction using receiver operating characteristic (ROC) analysis (AUC, sensitivity, specificity, and predictive values). In our analysis, CRS3 (complete or near-complete response) is treated as the positive class, while CRS1 (minimal or no response) and CRS2 (partial response) are treated as the negative class. The operating threshold ($\tau_{\mathrm{opt}}$) was chosen to prioritize precision, defined as:

\begin{equation}
\text{Precision} = \frac{\text{True Positives (TP)}}{\text{True Positives (TP)} + \text{False Positives (FP)}}
\label{eq:precision}
\end{equation}

\begin{itemize}
    \item \textit{False Positives:} patients predicted as CRS3 but who are actually CRS1-2.
    \item \textit{False Negatives:} patients predicted as CRS1-2 but who are actually CRS3.
\end{itemize}

As an investigational, pre-treatment decision-support tool, we focus on precision to minimize false positives. A false positive prediction (predicting that a patient is a complete or near-complete responder when they are not) is more concerning than a false negative. This is because false positives may overestimate chemosensitivity and prognosis, which could lead to overly optimistic expectations in clinical decision-making. So, the threshold is set conservatively to reduce false positives, even if that slightly increases false negatives. The model output is presented to the multidisciplinary team (MDT) as evidence that a patient is a good responder, influencing confidence in ongoing therapy. Treatment decisions remain primarily guided by biomarkers, platinum response, and clinical factors rather than CRS alone, so the model output must be considered within this broader clinical context.

\section{Results}\label{sec:results}

\begin{table}[p]
  \centering
  \caption{ (a) Discriminative performance across external cohorts (ROC-AUC, F1-score, precision, recall, accuracy). (b) Corresponding confusion matrix components (percentages relative to positives/negatives: TP, TN, FP, FN). All values are medians with 95\% confidence intervals, evaluated at ($\tau_{0.5}$ and $\tau_{opt}=0.69$).}
  \label{tab:combined_performance_confusion}

  \setlength{\tabcolsep}{4pt}
  \begin{tabular*}{\textwidth}{@{\extracolsep\fill}l|ccccc}
    \toprule
    \multicolumn{6}{l}{\textbf{(a) Performance metrics}} \\
    \midrule
    \textbf{Cohort} & \textbf{ROC-AUC} & \textbf{F1-score}
      & \textbf{Precision} & \textbf{Recall} & \textbf{Accuracy} \\
    \midrule
    IEO $\tau_{0.5}$ & 0.95 [0.95--0.96] & 0.52 [0.50--0.53]
      & 0.35 [0.33--0.36] & 1.00 [1.00--1.00] & 0.78 [0.78--0.78] \\
    IEO $\tau_{opt}$ & 0.95 [0.95--0.96] & 0.80 [0.80--0.80]
      & 0.80 [0.80--0.83] & 0.80 [0.80--0.83] & 0.95 [0.95--0.95] \\
    \midrule
    OV04 $\tau_{0.5}$ & 0.68 [0.53--0.81] & 0.58 [0.40--0.72]
      & 0.54 [0.35--0.74] & 0.62 [0.42--0.81] & 0.64 [0.52--0.76] \\
    OV04 $\tau_{opt}$ & 0.68 [0.53--0.81] & 0.38 [0.15--0.59]
      & 0.75 [0.40--1.00] & 0.26 [0.09--0.46] & 0.67 [0.55--0.79] \\
  \end{tabular*}

  \vspace{2pt}

 \setlength{\tabcolsep}{8pt}
  \begin{tabular*}{\textwidth}{@{\extracolsep\fill}l|cccc}
    \toprule
    \multicolumn{5}{l}{\textbf{(b) Confusion matrix elements
      \% w.r.t.\ Positives and Negatives size}} \\
    \midrule
    \textbf{Cohort} & \textbf{True Positives}
      & \textbf{True Negatives}
      & \textbf{False Positives} & \textbf{False Negatives} \\
    \midrule
    IEO $\tau_{0.5}$ & 100 [100--100] & 75 [61--89]
      & 25.0 [11--39] & - \\
    IEO $\tau_{opt}$ & 80 [28--100]   & 97 [91--100]
      & 3 [0--9]      & 20 [0--66] \\
    \midrule
    OV04 $\tau_{0.5}$ & 61 [42--81] & 66 [50--81]
      & 34 [19--50] & 39 [19--58] \\
    OV04 $\tau_{opt}$ & 26 [9--45]  & 94 [86--100]
      & 6 [0--14]   & 74 [55--91] \\
    \bottomrule
  \end{tabular*}

  \vspace{1.5em}
  
  \subcaptionbox{\label{fig:panel-a}}[.48\linewidth]{%
    \includegraphics[width=\linewidth]{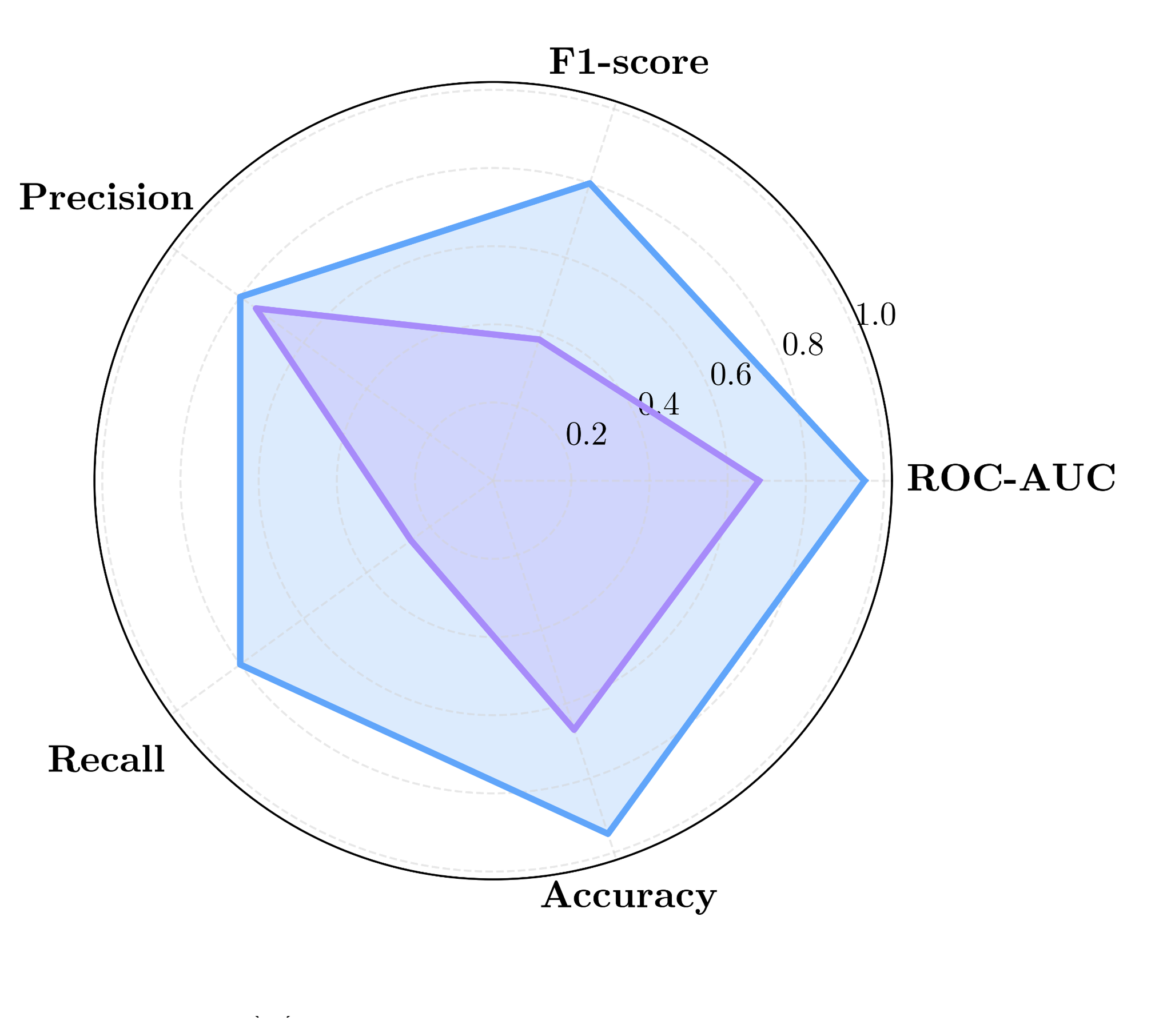}}
  \hfill
  \subcaptionbox{\label{fig:panel-b}}[.48\linewidth]{%
    \includegraphics[width=\linewidth]{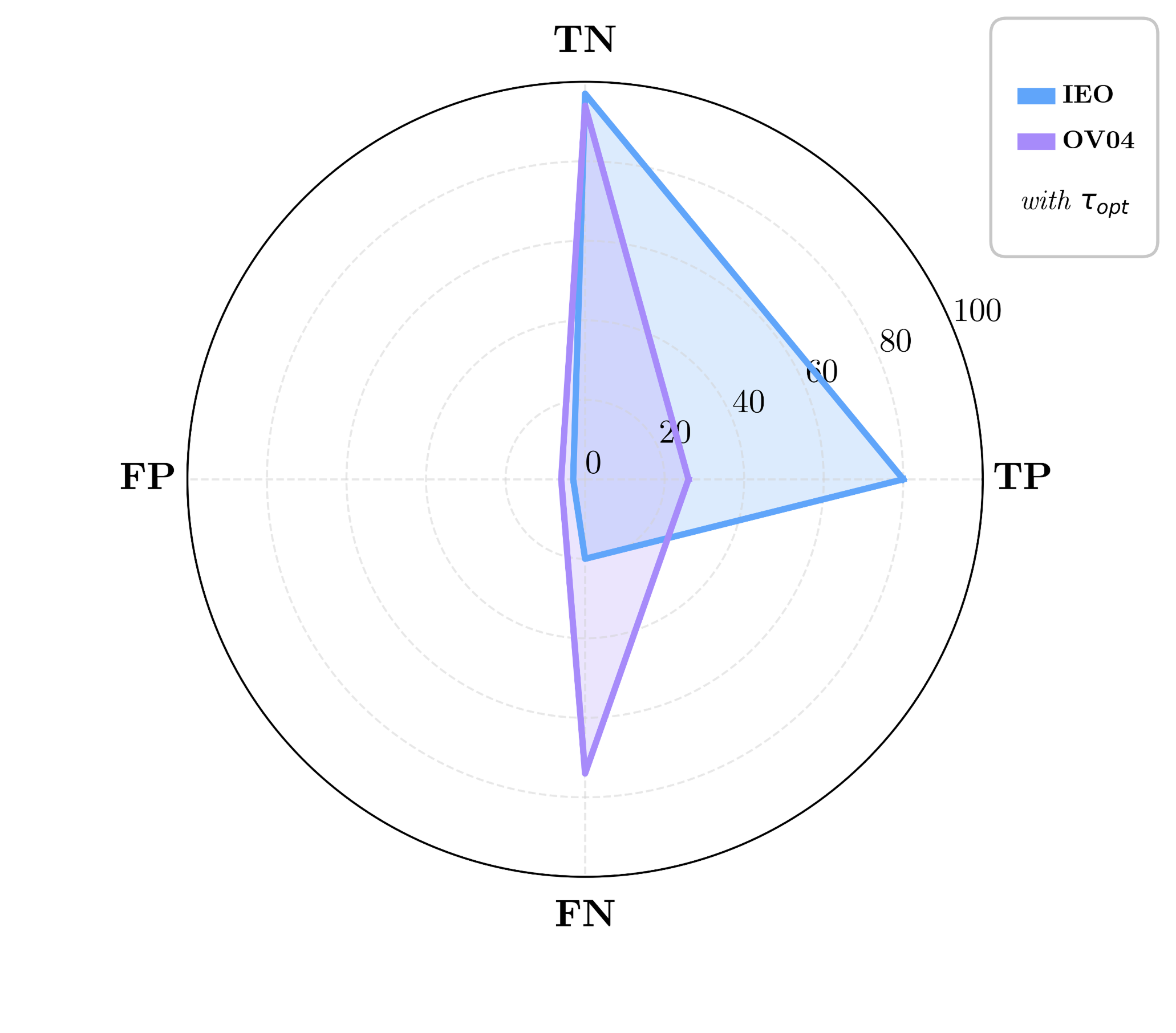}}

  \vspace{0.5em}

  \subcaptionbox{\label{fig:panel-c}}[.48\linewidth]{%
    \includegraphics[width=\linewidth]{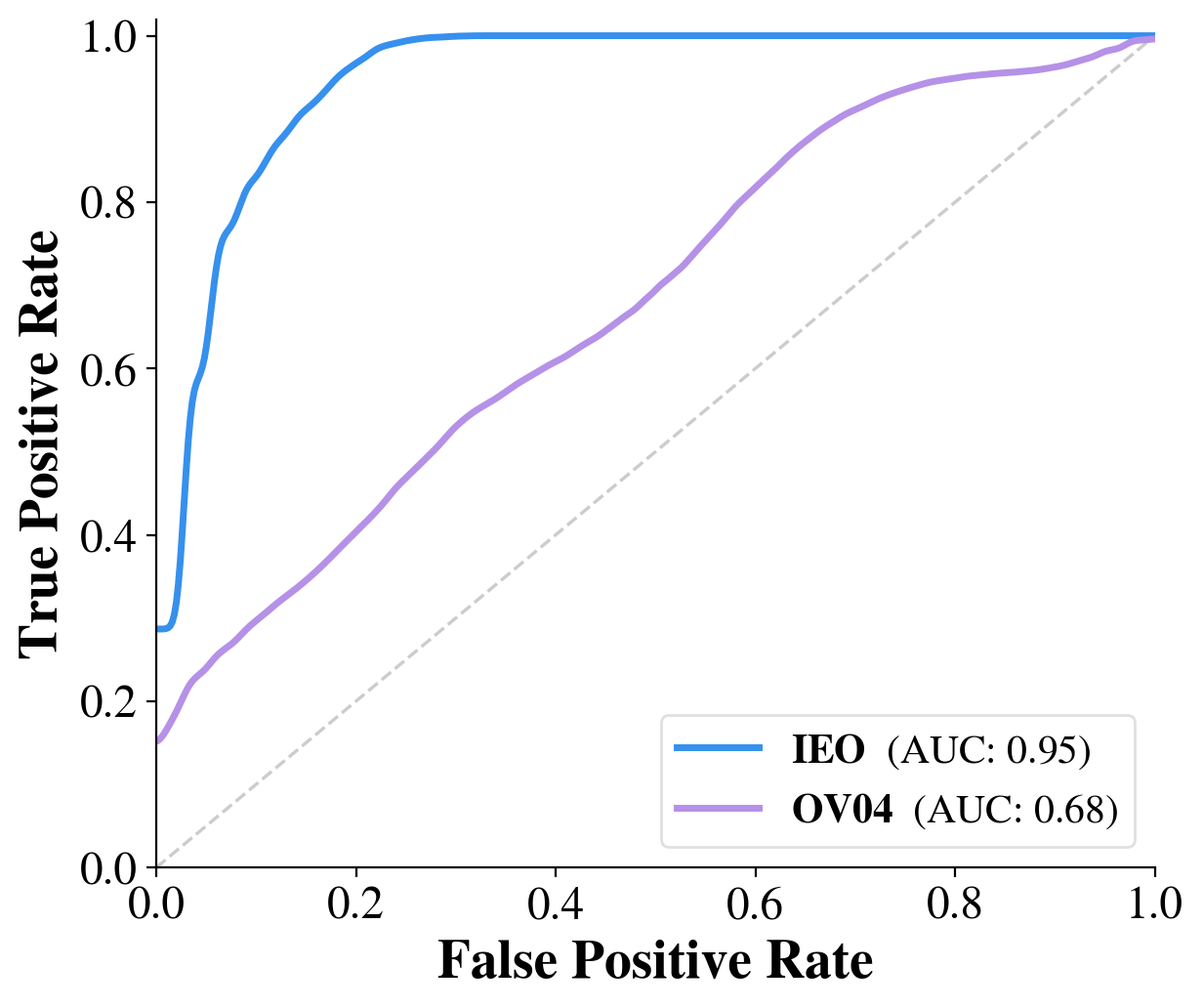}}
  \hfill
  \subcaptionbox{\label{fig:panel-d}}[.48\linewidth]{%
    \includegraphics[width=\linewidth]{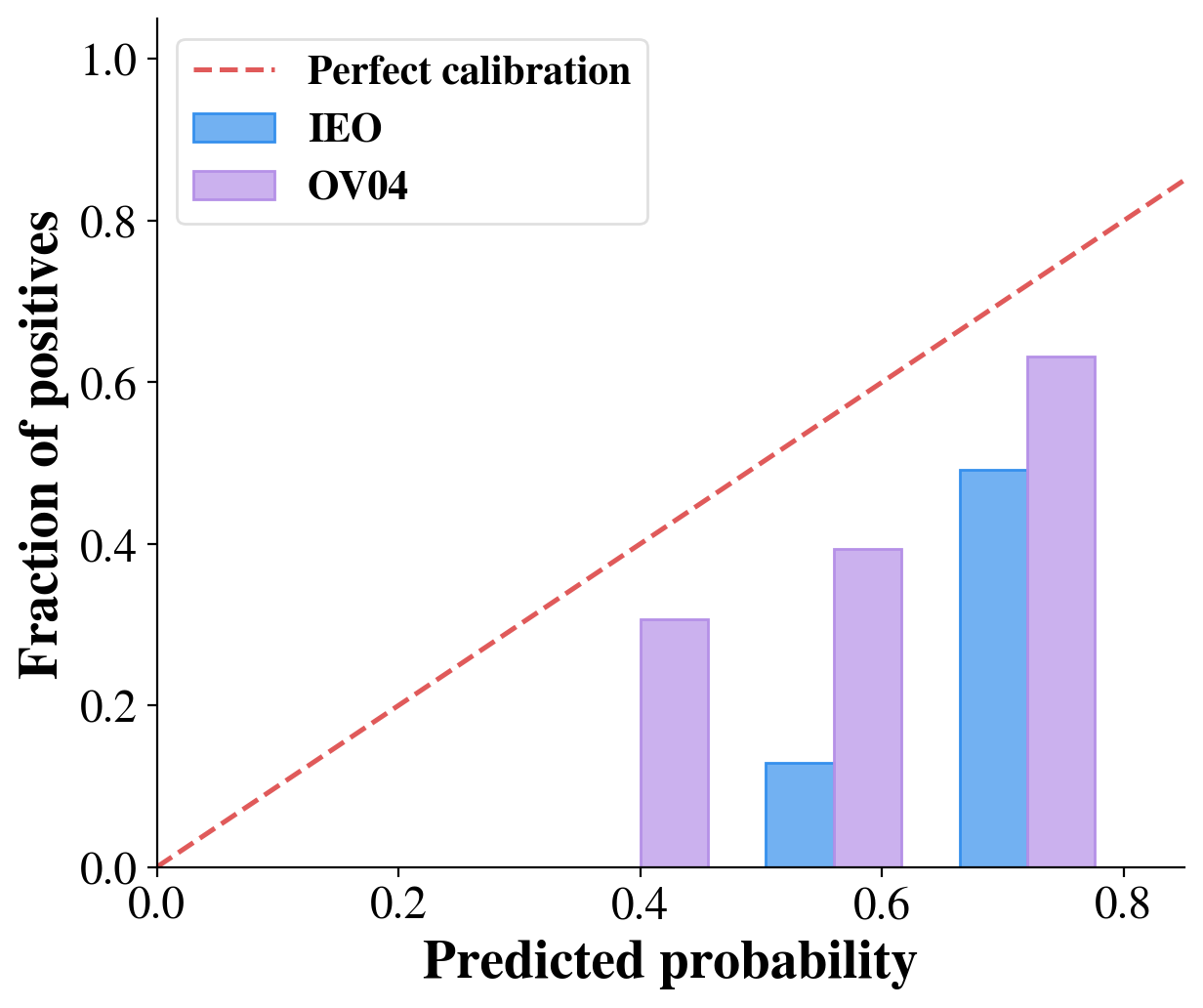}}
    \captionof{figure}{%
    {Classification performance (CRS1-2 vs.\ CRS3) on the IEO internal test set (threshold = 0.69) and OV04 external test set (threshold = 0.43). \textbf{(a)} ROC-AUC, F1, Precision, Recall, and Accuracy. \textbf{(b)} Confusion matrix elements (TP, TN, FP, FN). \textbf{(c)} ROC curves. \textbf{(d)} Reliability diagram, comparing the actual distribution of positives with the observed fraction of positives; the dashed diagonal indicates perfect calibration.}}
  \label{fig:combined_results}
\end{table}

\subsection{Model performance for response prediction}
Full results are reported in Table~\ref{tab:combined_performance_confusion} and Figure~\ref{fig:combined_results}. Overall model performance on both internal and external datasets is summarized below. On the internal test set (IEO), the model demonstrated high discriminative capacity, achieving an ROC-AUC of 0.95 [0.95–0.96]. At the default threshold of 0.50, the accuracy was 0.78 and precision 0.36. Optimizing the threshold on the pooled training data resulted in an optimal value of 0.69, substantially improving accuracy to 0.95 and precision to 0.80. On the external validation set (OV04), using the transferred optimal threshold ($\tau_{opt} = 0.69$), the accuracy is 0.67, and the corresponding precision is 0.75.

The use of $\tau_{\mathrm{opt}}$ resulted in a substantial reduction in false positive rates across both cohorts (from 25\% to 3\% in IEO; from 34\% to 6\% in OV04), with a corresponding increase in precision (from 0.35 to 0.80 in IEO; from 0.54 to 0.75 in OV04). However, this improvement was accompanied by an increase in false negatives, reflecting a systematic reduction in sensitivity. This trade-off underscores the inherent tension between minimizing false positives and avoiding the under-detection of true responders.

\subsection{Multimodal integration evaluation}
\begin{figure}[H]
    \centering

    \begin{subfigure}{0.48\linewidth}
        \centering
        \includegraphics[width=\linewidth]{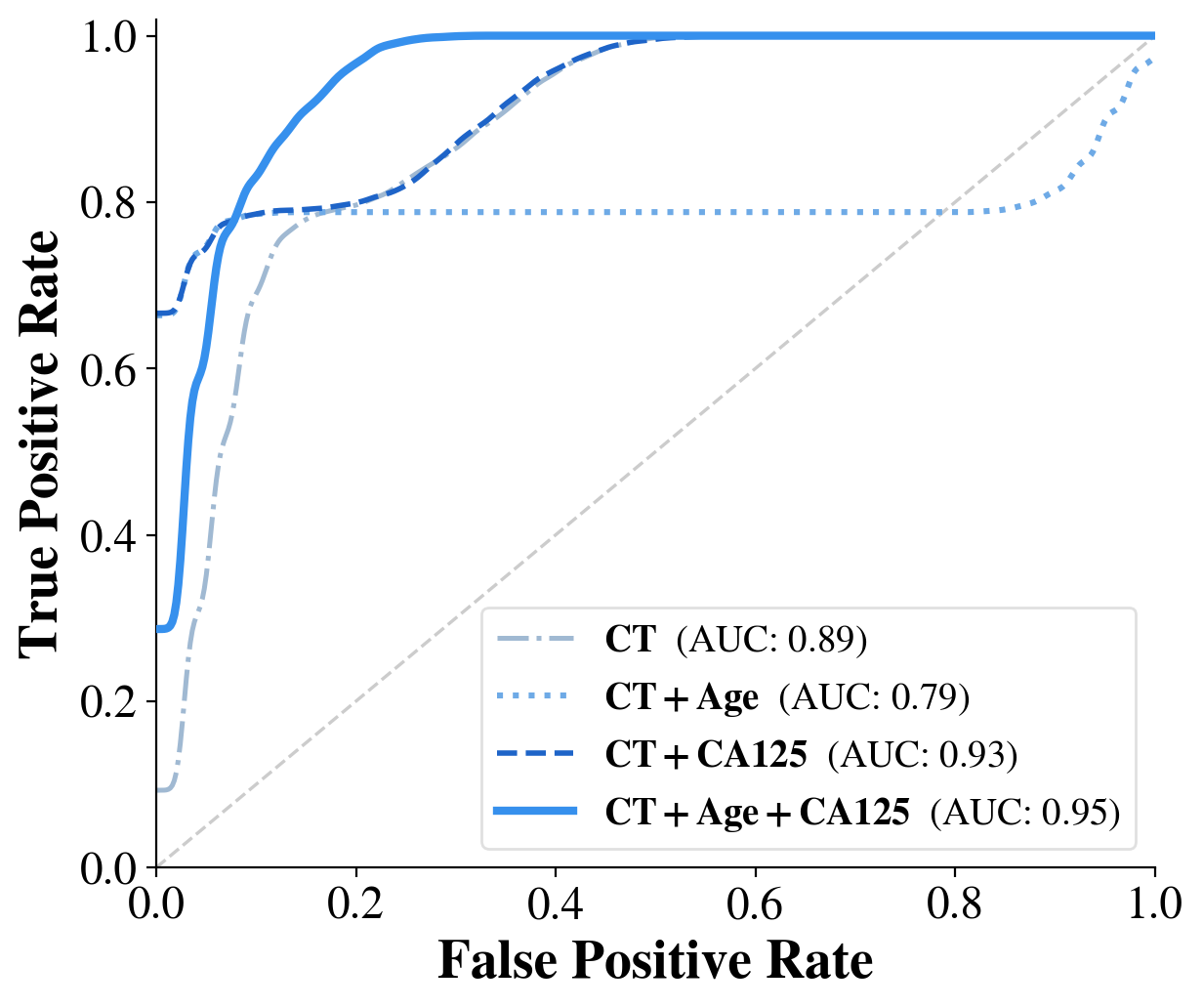}
    \end{subfigure}
    \hfill
    \begin{subfigure}{0.48\linewidth}
        \centering
        \includegraphics[width=\linewidth]{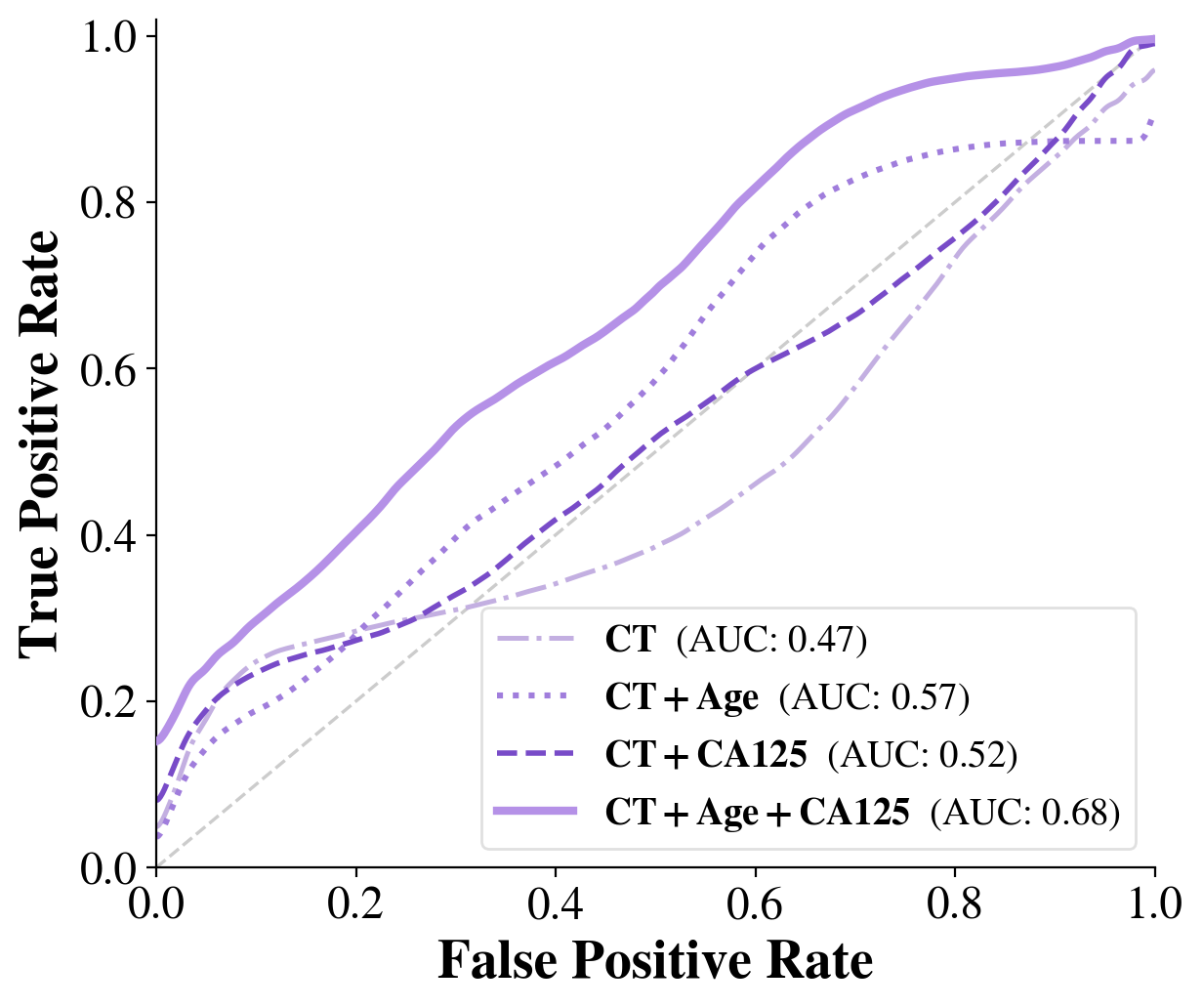}
    \end{subfigure}

    \caption{ROC curves for the ablation study on \textbf{(a)} IEO and \textbf{(b)} OV04 cohorts. The full multimodal model (CT + Age + CA-125) achieves the highest AUC, with the largest gains on OV04 observed when clinical features are added.}
    \label{fig:ablation_rocs}
\end{figure}

To evaluate the contribution of multimodal data integration, an ablation study was conducted across progressively enriched feature sets, including: (i) ViT-derived imaging features alone, (ii) imaging combined with age, (iii) imaging combined with CA-125, and (iv) the full multimodal configuration. As a baseline, a standalone random forest model was trained using only clinical and morphological variables (age, CA-125, and lesion volume). As shown in Table~\ref{tab:ablation} and Figure~\ref{fig:ablation_rocs}, incorporating an additional modality improved AUC incrementally across both the internal IEO test split and the external OV04 cohort, with the fully integrated model achieving peak performance in both settings.

\begin{table}[htbp]
\centering
\caption{Median AUC (95\% CI) for each feature configuration in the ablation study, evaluated on the IEO test split and the external OV04 cohort. 
Each row indicates which features are included: CT imaging, lesion volume, age, and CA-125 levels (\ding{51} = included, \ding{55} = excluded). 
AUC values are reported at the threshold maximizing the F1 score for each cohort.}
    \label{tab:ablation}
    \begin{tabular*}{\textwidth}{cccc@{\extracolsep{\fill}} cc}
        \toprule
        & & & & \multicolumn{1}{c }{IEO Test} & \multicolumn{1}{c}{OV04} \\
        CT & Volume & Age & CA125 & AUC & AUC \\
        \midrule
        \ding{55}& \ding{51} & \ding{51} & \ding{51} & 0.644 [0.459--0.792] & 0.537 [0.407--0.674] \\
        \ding{51}& \ding{55} & \ding{55} & \ding{55} & 0.894 [0.727--1.000] & 0.469 [0.312--0.632] \\
        \ding{51}& \ding{55} & \ding{51} & \ding{55} & 0.811 [0.333--1.000] & 0.572 [0.414--0.727] \\
        \ding{51}& \ding{55} & \ding{55} & \ding{51} & 0.933 [0.744--1.000] & 0.521 [0.364--0.677] \\
        \ding{51}& \ding{55} & \ding{51} & \ding{51} & \textbf{0.952 [0.862--1.000]} & \textbf{0.677 [0.527--0.806]} \\
        \bottomrule
    \end{tabular*}
\end{table}

\newpage The integration of CT-derived features with clinical variables improved predictive performance. While the imaging-only model demonstrated strong discrimination on the internal IEO cohort (AUC = 0.894), its performance dropped substantially on the external OV04 dataset (AUC = 0.469), indicating limited generalizability. In contrast, the full multimodal model achieved superior performance on both cohorts (AUC = 0.952 for IEO; AUC = 0.677 for OV04), highlighting the complementary contribution of clinical features in mitigating domain shift. Tumour morphology demonstrates consistent response-related trends across both the IEO (internal) and OV04 (external) cohorts, with decreasing tumour volume, surface area, and largest connected component fraction observed from CRS1 to CRS3. However, systematic differences are evident between datasets. For instance, morphological characteristics reveal differences in pre-treatment tumor volume distributions for CRS3 between cohorts, as reported in Table~\ref{tab:baseline_characteristics}.

\section{Discussion}\label{sec:conclusion}

We propose a multimodal integration framework that combines imaging and clinical variables to predict pathological complete response in HGSOC, aiming to improve generalizability across heterogeneous clinical settings. The model demonstrated strong internal performance on the IEO cohort, achieving a ROC-AUC of 0.95, balanced precision of 0.80, and maintained encouraging performance when evaluated on the external OV04 cohort (ROC-AUC 0.68 and balanced precision 0.75), supporting its potential for cross-institutional applicability. The ablation study clarifies the contribution of different data modalities: while imaging features alone capture relevant morphological patterns, the inclusion of clinical metadata, specifically age and CA125, improves external AUC by approximately 20\%. This highlights the complementary role of clinical variables, which appear to act as stabilizing factors, supporting more consistent behavior when the model is applied to new datasets.

Differences in tumor morphology between cohorts likely contribute to the observed variation in performance. In particular, variations in tumor volume distributions and morphological characteristics associated with CRS1-2 and CRS3 were observed between IEO and OV04. The model likely captures associations between tumor burden, surface-area-to-volume characteristics, and treatment response; however, shifts in these distributions across cohorts may influence how these learned patterns are expressed in external data. In this context, the addition of clinical variables appears to mitigate the impact of such morphological variability, reducing reliance on imaging features alone and supporting more consistent generalization across settings.

Taken together, these factors highlight the challenges of applying predictive models across heterogeneous clinical environments, while also supporting the role of multimodal approaches. Future work should focus on expanding multi-center training datasets, exploring domain adaptation strategies, and improving standardization in both clinical and pathological to ensure consistent performance across different institutions.

Another promising direction involves incorporating features from post-treatment assessments. 
Dynamic biomarkers such as CA-125 levels at the time of interval debulking surgery and the CA-125 KELIM index, which reflects the rate of CA-125 decline during chemotherapy, may provide additional predictive value beyond pre-treatment variables and could be integrated into future sequential modeling frameworks. 
Additionally, recent work has highlighted that adnexal and ovarian lesions may exhibit response patterns distinct from omental disease \cite{santoro2022prognostic}, suggesting that extending the imaging analysis to include ovarian and adnexal tumor features could further enrich the model's representation of treatment response. 
Both directions represent natural extensions of the present framework and provide promising avenues for further development of the model.

\section{Data and code availability}
The data supporting the findings of this study are not publicly available due to ethical regulations. To ensure reproducibility, the code for the model architecture, training, and inference is openly available on GitHub: https://github.com/FrancescaFati/UnderXAI-OVIT.

\section{Acknowledgments}
\label{sec:acknowledgments}
We acknowledge funding and support from Cancer Research UK (A22905) and the Cancer Research UK Cambridge Centre [CTRQQR-2021-100012 and A25177], The Mark Foundation for Cancer Research [RG95043]. This work was funded in partnership by Wellbeing of Women and the British Gynaecological Cancer Society (Award Ref: PRF608). Additional support was provided by the National Institute for Health Research (NIHR) Cambridge Biomedical Research Centre [NIHR203312 and BRC-1215-20014]. This work was supported by the project Under-XAI: understanding ovarian cancer initiation and progression through explainable AI, funded by the National Recovery and Resilience Plan (PNRR) under the EU’s NextGenerationEU programme (Project code: PNRR-MAD-2022-12376574; CUP: J47G22000530001). This study was also funded by the European Union’s Horizon Europe programme under Grant Agreement 101057389 (CINDERELLA project).

\section{Compliance with ethical standards}
This research was conducted retrospectively using de-identified patient data and in accordance with the principles of the Declaration of Helsinki. Approval for the use of private research data was granted by the Human Biology Research Ethics Committee of the University of Cambridge on 25 April 2023. The study was also approved by the Scientific Board of the European Institute of Oncology under protocol UID 4134. All patients provided informed consent for the use of their data for research purposes.

\bibliography{sn-bibliography}

\end{document}